%% file: main.tex
\documentclass[10pt,twocolumn,letterpaper]{article}

\usepackage{cvpr}
\usepackage{times}
\usepackage{epsfig}
\usepackage{graphicx}
\usepackage{amsmath}
\usepackage{amssymb}
\usepackage{cite}
\usepackage{subcaption}

\newcommand{\modelname}{RetinaTrack\,}

\newcommand{\Conv}{\mbox{Conv}}
\newcommand{\Sigmoid}{\mbox{Sigmoid}}

\newcommand{\denselist}{
\itemsep -2pt\topsep-8pt\partopsep-8pt
}

\PassOptionsToPackage{hyphens}{url}\usepackage[pagebackref=true,breaklinks=true,letterpaper=true,colorlinks,bookmarks=false]{hyperref}

\cvprfinalcopy %

\def\cvprPaperID{10568} %

\makeatletter
\newcommand{\printfnsymbol}[1]{%
  \textsuperscript{\@fnsymbol{#1}}%
}
\makeatother

\ifcvprfinal\fi
\begin{document}

\title{RetinaTrack: Online Single Stage Joint Detection and Tracking
}

\author{Zhichao Lu\thanks{Equal contribution with names listed alphabetically.}\,\,,\,
Vivek Rathod\printfnsymbol{1},\, Ronny Votel,\, Jonathan Huang \\
Google  \\
{\tt\small {\{lzc,rathodv,ronnyvotel,jonathanhuang\}}@google.com}
}

\maketitle

\begin{abstract}\vspace{-3mm} Traditionally multi-object tracking and object detection are performed using separate systems with most prior works focusing exclusively on one of these aspects over the other.  Tracking systems clearly benefit from having access to accurate detections, however and
there is ample evidence in literature that detectors can benefit from tracking
which, for example, can help to smooth predictions over time.  
In this paper we focus on the tracking-by-detection paradigm for  autonomous driving  where both tasks are mission critical.  We propose a conceptually simple and efficient joint model of detection and tracking, called \modelname, which modifies the popular single stage RetinaNet approach such that it is
amenable to instance-level embedding training. We show, via evaluations on the Waymo Open Dataset, that we outperform a recent state of the art tracking algorithm while requiring significantly less computation. We believe that our simple yet effective approach can serve as a strong baseline for future work in this area.
\end{abstract}\vspace{-4mm}

\vspace{-2mm}
\section{Introduction}\label{sec:intro}
\vspace{-1mm}
\input{introduction}

\vspace{-2mm}
\section{Related Work}\label{sec:related}
\vspace{-1mm}

\input{relatedwork}

\vspace{-2mm}
\section{The \modelname Architecture}\label{sec:model}
\vspace{-1mm}

\input{modelarchitecture}

\vspace{-2mm}
\section{Experiments}\label{sec:experiments}
\vspace{-2mm}

\input{experiments}

\vspace{-2mm}
\section{Conclusion}\label{sec:conclusion}
\vspace{-1mm}

\input{conclusion}

\subsubsection*{Acknowledgements}\vspace{-2mm}
{\small
We are grateful to Sara Beery,
 Yuning Chai,
 Wayne Hung,
Henrik Kretzschmar,
David Ross, 
Tina Tian, and Jack Valmadre
  for valuable discussions.
}

{\small
\bibliographystyle{ieee_fullname}
\bibliography{references}
}

\end{document}

%% file: introduction.tex
\thispagestyle{empty}
The tracking-by-detection paradigm today has become the dominant method for multi object tracking (MOT) and works by detecting objects in each frame independently and then performing data association across frames of a video.  In recent years, both aspects of this approach (detection and data association) have  seen significant technological advances due to the adoption of deep learning.  

Despite the fact that these two tasks often go hand in hand and the fact that deep learning has made models easily amenable to multitask training, even today  it is far more common to separate the two aspects than to train them jointly in one model, with most papers often focusing on detection metrics or tracking metrics and rarely both.  This task separation has led to more complex models and less efficient approaches.  It is telling that the flagship benchmark in this area (MOT Challenge~\cite{milan2016mot16}) assumes that models will make use of publicly available detections and that papers continue to claim the use of a real-time tracker while not measuring the time required to perform detection.

\begin{figure*}[t!]
\centering
\includegraphics[width=\linewidth]{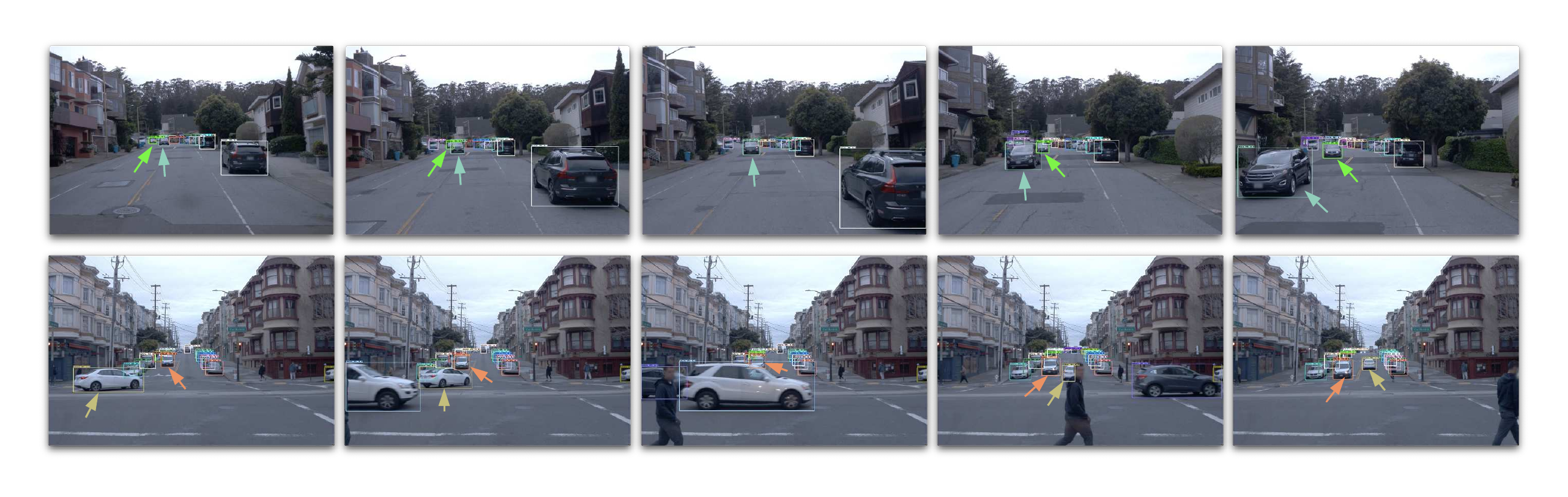}\vspace{-6mm}
\caption{\footnotesize Example vehicle tracking results on the Waymo Open Dataset --- tracks
are color coded and for clarity we highlight two tracks 
in each sequence with arrows.
Challenges in this dataset include small objects, frequent occlusions
due to other traffic or pedestrians, changing scales and low illumination.}
\label{fig:prettypictures}
\end{figure*}

In this paper we are interested primarily in the autonomous driving domain where object detection and multi-object tracking are mission-critical technologies.  If we cannot detect and track, we will not be able to predict where vehicles and pedestrians are going (and at what speed), and consequently we will not, e.g., know whether to yield to a pedestrian at the corner or whether to drive full speed down a street despite cars coming down an opposing lane. 

We focus specifically on RGB inputs which, while typically not the only sensing modality used within a modern autonomous vehicle, play an important role; RGB cameras do not have the same range constraints as LIDAR, are considerably cheaper and are capable of detecting much smaller objects which are important particularly for highway driving where the faster driving speeds make it important to be able to react to distant vehicles or pedestrians.

In the setting of autonomous driving, speed and accuracy are both essential and therefore the choice of  architecture is critical as one cannot simply take the heaviest/most performant model or the most lightweight but not as accurate model.  
We base our model on the RetinaNet detector~\cite{lin2017focal} which is real-time while reaching state of art accuracy and is specifically designed to detect small objects well.  To this base detector, we add instance-level embeddings for the purposes of data association.  However the vanilla RetinaNet architecture is not suitable for these per-instance embeddings --- we propose a simple but effective modification to RetinaNet's post-FPN prediction 
subnetworks to address these issues.  We show via  ablations  that our model, which we dub, \emph{\modelname}, benefits from joint training 
of the tracker and detector. It has small computational overhead compared to base RetinaNet and is therefore fast ---  due to its simplicity, it is also easy to train via Google TPUs.

To summarize, our main contributions are as follows:
\begin{itemize}\denselist
\item We propose a jointly trained detection and tracking model - our method is  simple, efficient and could be feasibly deployed in an autonomous vehicle.
\item We propose a simple modification to single shot detection architectures that allow for extracting instance level features; we use these features for tracking, but they could also be useful for other purposes.
\item We establish initial strong baselines for detection and tracking from 2d images on the Waymo Open dataset~\cite{waymo_open_dataset} (Figure~\ref{fig:prettypictures}) and show that our method achieves state of the art performance.
\end{itemize}\vspace{-1mm}
We hope our simple model will serve as a solid baseline and ease future research in joint detection and tracking.

%% file: relatedwork.tex
Traditionally multi-object tracking and detection have been treated in two separate literatures with trackers often using detectors as black box modules but not necessarily incorporating them deeply.  In recent years both fields have begun to rely heavily on deep learning which makes it natural to model both tasks jointly.  However with a few exceptions, joint training of detection and tracking remains the exception rather than the rule.  And there are few papers that evaluate both tracking \emph{and} detection with papers often focusing on one evaluation exclusively.

\vspace{-1mm}
\subsection{Object Detection in Images and Video}
\vspace{-1mm}

In recent years there has been an explosion of technological progress in the field of object detection driven largely by community benchmarks like the COCO challenge~\cite{lin2014microsoft} and Open Images \cite{kuznetsova2018open}.  There have also been a number of advances in detection specific model architectures including anchor-based models, both single stage (e.g., SSD \cite{liu2016ssd}, RetinaNet \cite{lin2017focal}, Yolo variants \cite{redmon2016you,redmon2017yolo9000}) and two-stage detectors (e.g., Fast/Faster R-CNN~\cite{girshick2015fast,ren2015faster,he2016deep}, R-FCN~\cite{dai2016r}), as well as the newer anchor-free models (e.g., CornerNet~\cite{law2018cornernet,law2019cornernet}, CenterNet~\cite{zhou2019objects}, FCOS~\cite{tian2019fcos}).

Building on these single frame architectures are methods  incorporating temporal context for better detection in video (specifically to combat  motion blur, occlusion, rare poses of objects, etc).  Approaches include the use of 3d convolutions (e.g., I3D, S3D)~\cite{carreira2017quo,luo2018fast,xie2018rethinking} or recurrent networks~\cite{liu2018mobile,kang2017object} to extract better temporal features.  There are also 
a number of works that use tracking-like concepts of some form in order to aggregate, but their main focus lies in detection and not tracking.  For example, there are works  exploiting flow (or flow-like quantities) to aggregate features~\cite{zhu2017deep,zhu2017flow,zhu2018towards,bertasius2018object}.  Recently there are also papers that propose object level attention-based aggregation methods~\cite{wu2019long,shvets2019leveraging,wu2019sequence,deng2019object} which effectively can be viewed at high level as methods to aggregate features along tracks. In many of these cases, simple heuristics to ``smooth'' predictions along time are also used, including tubelet smoothing~\cite{gkioxari2015finding} or SeqNMS~\cite{han2016seq}.

\vspace{-1mm}
\subsection{Tracking}
\vspace{-1mm}

Traditionally trackers have played several different roles.  In the cases mentioned above, the role of the tracker has been to improve detection accuracy in videos (e.g. by smoothing predictions over time).   In other cases, trackers have also been used to augment (traditionally much slower) detectors allowing for real-time updates based on intermittent detector updates (e.g. \cite{bouguet2001pyramidal,benfold2011stable}). 

Finally in applications such as self-driving and sports analysis, track outputs are themselves of independent interest.  For example typical behavior prediction modules take object trajectories as input in order to forecast future trajectories (and thereby react to) predicted behavior of a particular object (like a car or pedestrian)~\cite{urmson2008autonomous,chai2019multipath,tang2019multiple,yeh2019diverse,zhan2018generating}. 
In this role, the \emph{tracking-by-detection} paradigm has become the predominant approach taken for multi-object tracking, where detection is first run on each frame of an input sequence, then the results linked across frames (this second step is called \emph{data association}).

In the pre-deep learning era, tracking-by-detection methods~\cite{hamid2015joint,choi2015near,xiang2015learning} tended to focus on using whatever visual features were available and finding a way to combat the combinatorial explosion of the various graph optimization problems~\cite{reid1979algorithm,cox1996efficient,fortmann1983sonar} that have been formulated to determine optimal trajectories. In recent years, this trend has been reversed with authors using simple matching algorithms (e.g. Hungarian matching~\cite{munkres1957algorithms}) and focusing on learning features that are better for data association e.g, via deep metric learning~\cite{leal2016learning,son2017multi,schulter2017deep,wojke2017simple,tang2016multi,sadeghian2017tracking,bergmann2019tracking}. For example \cite{wojke2017simple} proposed Deep Sort, a simple yet strong baseline that takes offline detections (produced by Faster RCNN) and links them using an offline trained deep ReID model and Kalman filter motion model. In this context, our work can be viewed as a simplified pipeline compared to Deep Sort, 
relying on a more lightweight
detection network which is unified with a subnetwork tasked with performing ReID.

\vspace{-1mm}
\subsection{Detection meets Tracking}
\vspace{-1mm}

Strong  detection is critical to strong tracking. This can be seen via the commonly used CLEAR MOT metric (MOTA, multiple object tracking accuracy)~\cite{milan2016mot16} which penalizes false positives, false negatives and identity switches (the first two terms of which are detection related). The recent \emph{Tracktor} paper~\cite{bergmann2019tracking} pushes this observation to the limit achieving strong results using only a single frame Faster R-CNN detection model. Tracking itself is accomplished by exploiting the behavior of the second stage of Faster R-CNN that allows an imprecisely specified proposal (e.g. a detection from the previous frame) to be ``snapped'' onto the closest object in the image.  With a few small modifications (including an offline trained ReID component),  Tracktor  is currently state of the art on the MOT17 Challenge and we compare against this strong baseline in our experiments.

To address the issue that detection can have such an outsized impact on tracking metrics, benchmarks such as the MOT Challenge have tried to make things ``fair'' by having multiple methods use exactly the same out-of-the-box provided detections.  However this restriction unnecessarily ties ones hands as it assumes that the two will be done separately and consequently 
can preclude jointly trained models such as our own.  One wonders whether the paucity of joint detection/tracking literature may be due in part to this emphasis on using black box detections.  

Prior to our work, there have been several recent attempts to train joint tracking/detection models.  Feichtenhofer et al.~\cite{feichtenhofer2017detect} run an R-FCN (\cite{dai2016r}) base detection architecture  and simultaneously compute correlation maps between high level feature maps of consecutive frames which are then passed to a secondary prediction tower in order to predict frame-to-frame instance motion. Like~\cite{feichtenhofer2017detect}, we train for both tasks jointly. However where they focus exclusively on detection metrics for Imagenet Vid, motivated by autonomous driving needs, we evaluate both tracking and detection metrics.  Our architecture is also considerably simpler, faster and based on a stronger single stage detector.

There are also several works that predict 3d tubelets~\cite{hou2017tube,girdhar2018detect} directly using 3d inputs by using 2d anchor grids that are allowed to ``wiggle’’ in time via a predicted temporal sequence of spatial offsets.  However these methods are typically heavier and require a mechanism to associate tubelets amongst each other, often relying on simple heuristics combining single frame scores and IOU overlap.  We directly learn to associate detections (and show that this is useful).

Finally the work most related to our approach is Wang et al.~\cite{wang2019towards} which also combines an FPN based model (with YOLO v3) with an additional embedding layer. In contrast, we use a modification of RetinaNet which has stronger detection performance and we show that without our modifications to the FPN, performance suffers.

%% file: modelarchitecture.tex
\begin{figure*}
    \centering
    \begin{subfigure}[b]{0.32\linewidth}
        \centering
        \raisebox{10mm}{
        \includegraphics[width=.95\linewidth]{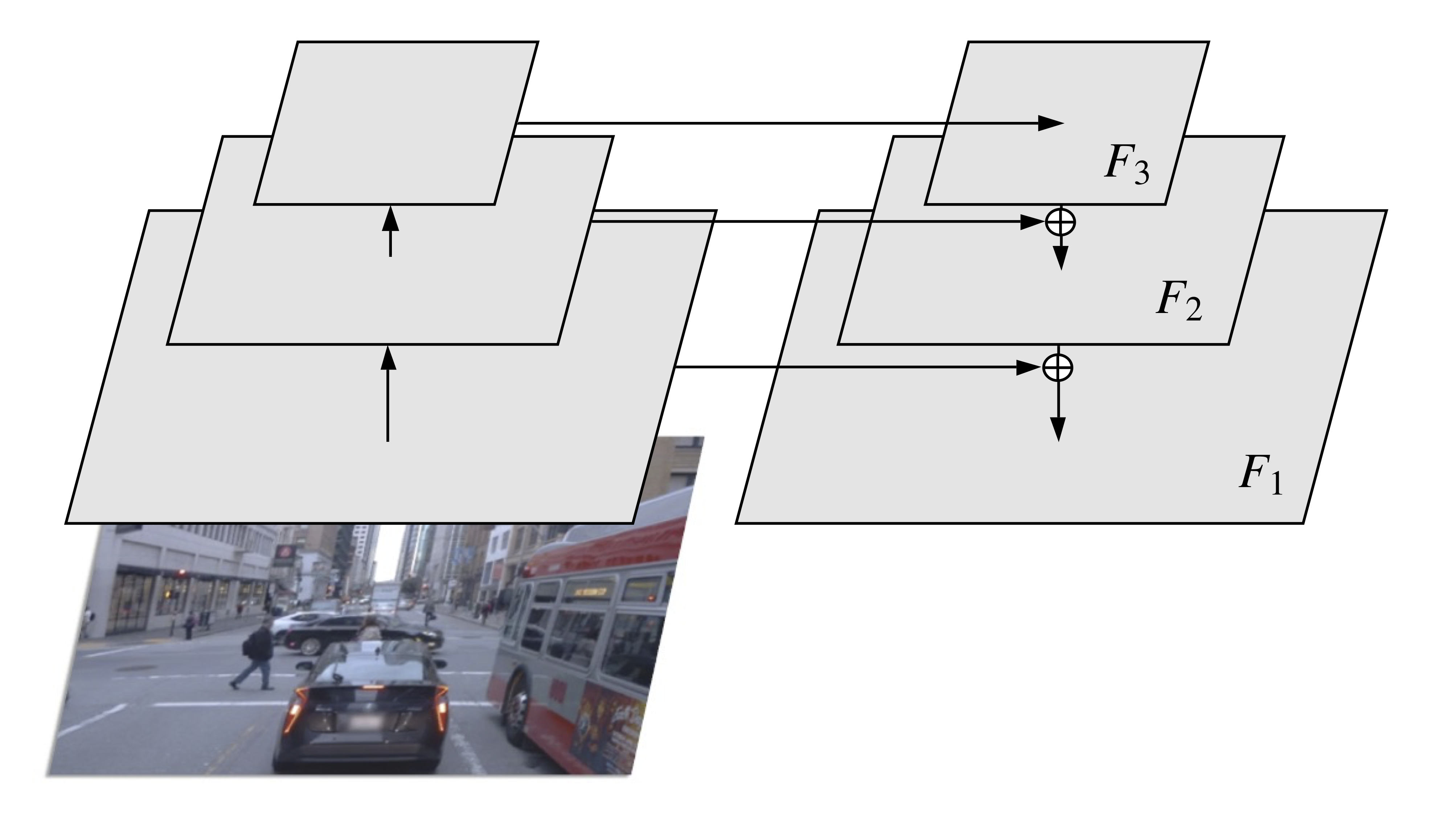}
        }\vspace{-4mm}
        \caption{}
        \label{fig:fpn}
    \end{subfigure}
    \begin{subfigure}[b]{0.28\linewidth}
        \centering
        \raisebox{0mm}{
        \includegraphics[width=.98\linewidth,trim=0 0 720 0, clip]{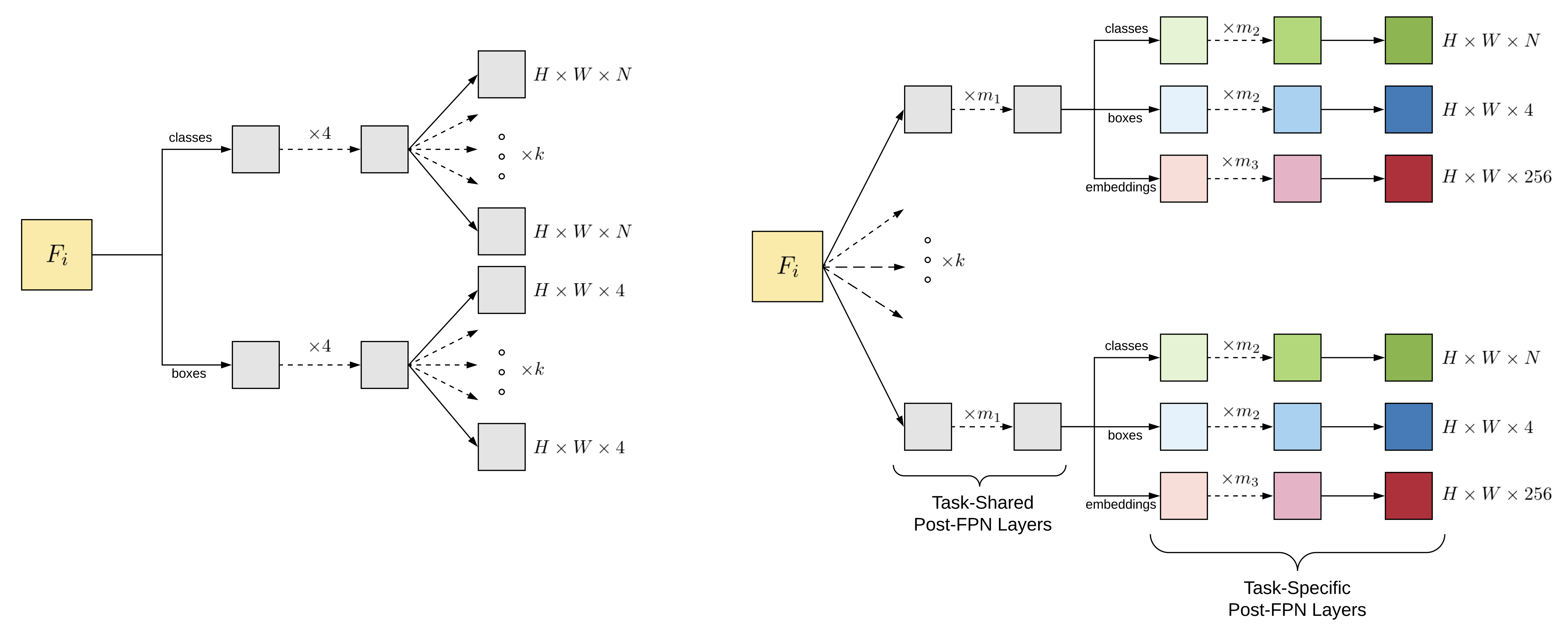}
        }\vspace{-5mm}
        \caption{}
        \label{fig:retinanet}
    \end{subfigure}
    \begin{subfigure}[b]{0.36\linewidth}
        \centering
        \raisebox{0mm}{
        \includegraphics[width=.98\linewidth,trim=550 0 0 0, clip]{figures/architectures.png}
        }\vspace{-3mm}
        \caption{} 
        \label{fig:tracktinanet}
    \end{subfigure}\vspace{-3mm}
    \caption{\footnotesize  \textbf{Architecture diagrams}. 
    (\subref{fig:fpn}) Feature Pyramid Network (FPN) and 
    Post-FPN layers of (vanilla)
    (\subref{fig:retinanet}) RetinaNet and (\subref{fig:tracktinanet})   \modelname.  In order to capture instance level features \modelname
    splits the computational pathways among different anchor shapes
    at an earlier stage in the Post-FPN subnetwork of RetinaNet.
    Yellow boxes $F_i$ represent feature
    maps produced by the FPN.  In both models we share convolutional parameters across all FPN layers.  At level of a single FPN layer, 
    gray boxes represent convolutional layers that are unshared while
    colored boxes represent sharing relationships (boxes with the same
    color share parameters).}
    \label{fig:architecture}
\end{figure*}

In this section we describe the design of a variant of RetinaNet that allows us to extract per-instance level features.  Like other anchor based detectors, every detection produced by RetinaNet is associated with an anchor.  In order to link a detection to those in another frame, we would like to be able to identify a feature vector associated with its corresponding anchor and pass it to an embedding network which will be trained with metric learning losses.

\begin{figure}
    \centering
        \includegraphics[width=.85\linewidth]{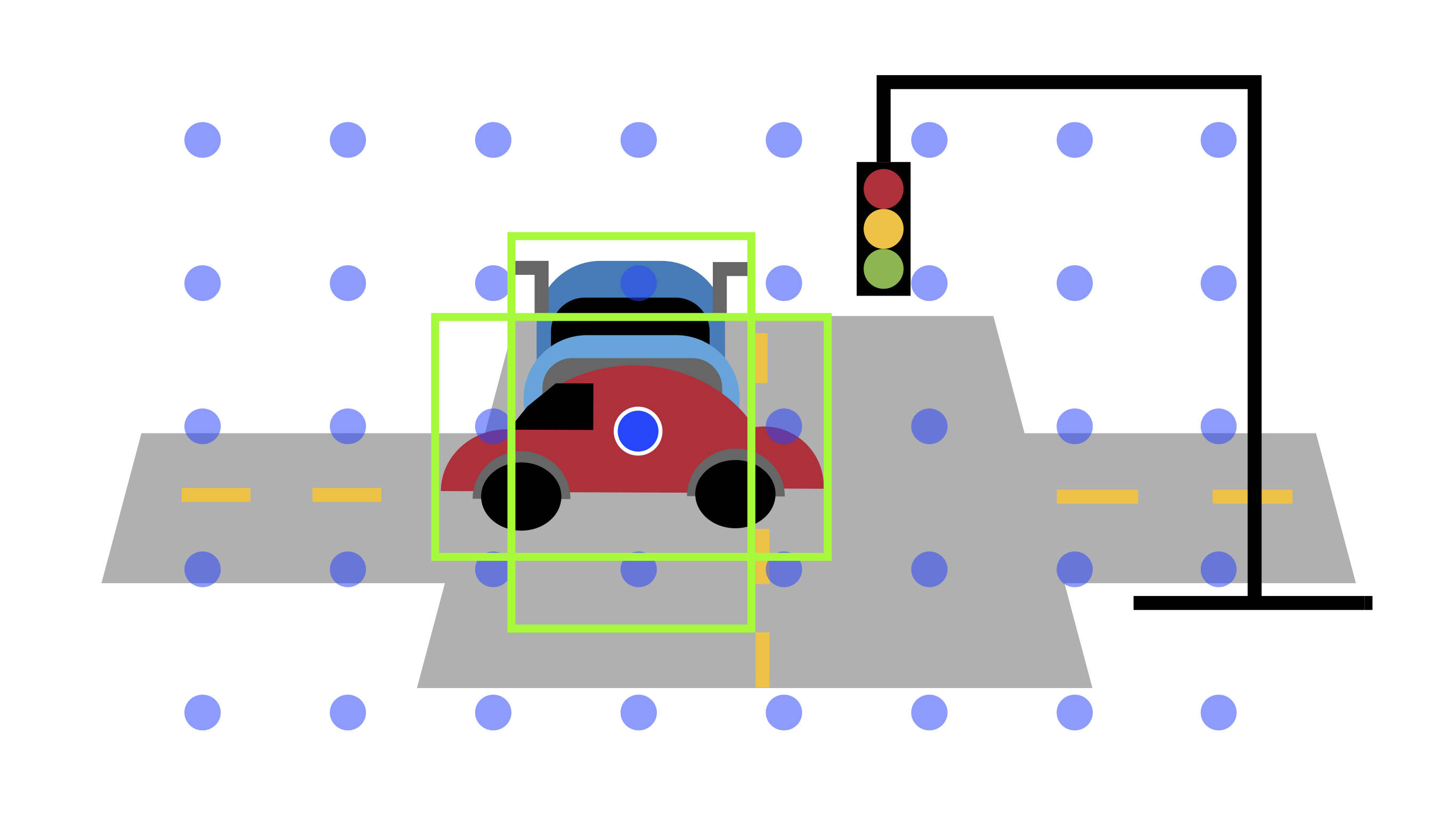}\vspace{-4mm}
    \caption{\footnotesize 
    In order to track successfully
    through occlusions, we need to be able to model that objects that share the same anchor grid
    center have distinct tracking features.  Here, green boxes 
    represent two anchors centered at the same location
    which match the cars in the scene. Blue dots represent 
    centers of the anchor grid.
    }
    \label{fig:traffic}\vspace{-4mm}
\end{figure}

\vspace{-1mm}
\subsection{RetinaNet}
\vspace{-1mm}

To begin, we review the popular RetinaNet
architecture~\cite{lin2017focal} and
explain why the vanilla model is not suitable for instance level embeddings.
Modern convolutional object detectors extract feature maps 
from sliding window positions arranged along a regular grid
over the image.  In anchor-based methods such as RetinaNet, we place
$K$ anchor boxes $\{A_1, \dots, A_K\}$ 
of different shapes (varying aspect ratios and sizes) on top of 
each grid point and ask the model to make predictions 
(e.g., classification logits, box regression offsets)
relative to these anchors.

In the case of RetinaNet, we use an FPN-based (feature pyramid network) feature 
extractor~\cite{lin2017feature} which produces multiple layers of
feature maps $F_i$ with different spatial resolutions $W_i\times H_i$
(Figure~\ref{fig:fpn}).  
Each feature map $F_i$ is then passed to two post-FPN task-specific convolutional subnetworks predicting $K$ tensors (one for
each possible anchor shape) $\{Y^{cls}_{i,k}\}_{k=1:K}$ each of shape 
$W_i\times H_i\times N$ representing $N$-dimensional 
classification logits, as well
as $K$ tensors
$\{Y^{loc}_{i,k}\}_{k=1:K}$ of shape
$W_i\times H_i\times 4$ representing box regression offsets.
(Figure~\ref{fig:retinanet}).  Note that typically papers collapse these outputs to be a
single combined tensor instead of $K$ tensors with one for each
anchor shape ---
however for our purposes we separate these predictions 
for clarity (with the end result being equivalent).

More formally, we can write the RetinaNet classification and location
prediction tensors as a 
function of each of the feature maps $F_i$ as follows:

\vspace{-3mm}
{\footnotesize
\begin{align}
Y^{cls}_{i,k}(F_i) &\equiv \Sigmoid(\Conv(\Conv^{(4)}(F_i;\,\theta^{cls});\,\phi^{cls}_k)),\label{eqn:retinanet_cls} \\
Y^{loc}_{i,k}(F_i) &\equiv \Conv(\Conv^{(4)}(F_i;\,\theta^{loc});\,\phi^{loc}_k),\label{eqn:retinanet_loc} 
\end{align}
}
\vspace{-3mm}

where $k\in\{1, \dots K\}$) indexes into the $K$ anchors.  We use 
$\Conv^{(4)}$ to refer to 4 intermediate $3\times 3$ convolution layers
(which include batch norm and ReLu layers unless otherwise
specified).  The model 
parameters after the FPN are $\theta^{cls}$, $\{\phi^{cls}_k\}_{k=1}^K$,
$\theta^{loc}$ and $\{\phi^{loc}_k\}_{k=1}^K$.  
Importantly, while the classification and box regression subnetworks
have different parameters for a given FPN layer,
the parameters are shared across FPN layers 
which allow us to treat  feature vectors extracted
from different layers as if they belonged to compatible embedding spaces.

\vspace{-1mm}
\subsection{Modifying task-prediction subnetworks to have anchor-level features}
\vspace{-1mm}

From Equations~\ref{eqn:retinanet_cls},\,\ref{eqn:retinanet_loc} all 
convolutional parameters of RetinaNet are shared
amongst all $K$ anchors until the final convolution for the classification
and regression subnetworks.
Therefore there is no clear way to extract per-instance features
since 
if two detections match to anchors at the
same location with different
shapes, then the only point in the network at which they would be distinguished are at the final class and box regression predictions.
This can be especially problematic when tracking through occlusions
when objects are more likely to correspond to anchors which 
share the same location (Figure~\ref{fig:traffic}).

Our solution is to force the split among the anchors to occur earlier
among the post-FPN prediction layers, 
allowing us to access intermediate level features
that can still be uniquely associated with an anchor (and consequently a final detection).  
Our proposed modification is simple --- we end up with a similar architecture to RetinaNet, 
but tie/untie weights in a different manner
compared to the vanilla architecture.  In our \modelname model we
predict via the following parameterization (c.f.
 Equations~\ref{eqn:retinanet_cls},\,\ref{eqn:retinanet_loc}):

\vspace{-2mm}
{\footnotesize
\begin{align}
F_{i,k} &= \Conv^{(m_1)}(F_i;\,\theta_k), \\
Y^{cls}_{i,k} &\equiv \Sigmoid(\Conv(\Conv^{(m_2)}(F_{i,k};\,\theta^{cls});\,\phi^{cls})), \\
Y^{loc}_{i,k} &\equiv \Conv(\Conv^{(m_2)}(F_{i,k};\,\theta^{loc});\,\phi^{loc}).
\end{align}
}
\vspace{-2mm}

Thus for each post-FPN layer $F_i$, we first apply $K$ 
convolutional sequences (with $m_1$ layers) in parallel to predict the $F_{i,k}$ tensors, which we view as \emph{per-anchor instance-level features},
since from this point on, there will be a unique $F_{i,k}$ associated with
every detection produced by the RetinaNet architecture
(Figure~\ref{fig:tracktinanet}).
We will refer to this first segment of the model as the
\emph{task-shared post-FPN layers} which use separate
parameters $\theta_k$ for each of the $K$ anchor shapes, but share 
$\theta_k$ across FPN layers (as well as the two tasks of classification
and localization).

The $F_{i,k}$ are not task-specific features, but we next apply two 
parallel sequences of \emph{task-specific post-FPN layers} to 
each $F_{i,k}$. 
Each sequence
consists of $m_2$ $3\times 3$ convolutions followed by a $3\times 3$  
final convolution with $N$ output channels in the case of classification 
logits (where $N$ is the
number of classes) and 4 output channels in the case of box regression offsets.
For our two task specific subnetworks, we share parameters 
$\theta^{cls}$, $\phi^{cls}$,
$\theta^{loc}$ and $\phi^{loc}$
across both the $K$ anchor shapes as well as all FPN layers so that after
the task-shared layers, all features can be considered as belonging to
compatible spaces.

\vspace{-1mm}
\subsection{Embedding architecture}
\vspace{-1mm}

Having the instance level features $F_{i,k}$ now in hand, we additionally
apply a third sequence of task-specific layers consisting of $m_3$ 
 $1\times 1$ convolution layers projecting the instance level features to a final track embedding space with each convolution layer mapping to 256 output channels:
\begin{equation}
Y^{emb}_{i,k} \equiv \Conv^{(m_3)}(F_{i,k};\,\theta^{emb}).
\end{equation}
We use batch norm~\cite{ioffe2015batch} and ReLU nonlinearities after each convolution except at the final embedding layer and use the same shared parameters across all FPN layers and all $K$ anchor shapes
(see again Figure~\ref{fig:tracktinanet}).

To summarize, \modelname predicts 
per-anchor instance-level features $F_{i,k}$.
Given a detection $d$, there is a unique anchor that
generated $d$ --- and the feature maps $F_{i,k}$ now give us a 
unique feature vector associated with $d$.
Where RetinaNet models run 4 convolutional layers for each of the two  task-specific subnetworks, in \modelname, each output tensor is the result
of $m_1+m_2+1$ (or $m_1+m_3$ in the case
of the track embeddings) convolutional layers where $m_1$,
$m_2$ and $m_3$ are
structural hyperparameters. We discuss ablations for these
settings further in Section~\ref{sec:experiments}.

\vspace{-1mm}
\subsection{Training details}\label{sec:training}
\vspace{-1mm}

At training time we minimize an unweighted sum of the two standard RetinaNet losses (Sigmoid Focal Loss for classification, and Huber Loss for box regression) as well as an additional embedding loss which encourages detections corresponding to the same track to have similar embeddings. Specifically we train with triplet loss~\cite{schroff2015facenet,chechik2010large} using the \emph{BatchHard} strategy for sampling triplets~\cite{hermans2017defense}.\vspace{-2mm}

{\footnotesize
\begin{equation}\label{eqn:batchhard}
\mathcal{L}_{BH}(\theta;\,X) = \sum_{j=1}^A \mbox{SoftPlus}\left(m + \max_{\substack{p=1\dots A \\ t_j=t_p}} D_{jp} 
        - \min_{\substack{\ell=1\dots A \\ t_j\neq t_\ell}} D_{j\ell}\right),
\end{equation}\vspace{-3mm}
}

\noindent where $A$ is the number of anchors that match to groundtruth boxes, $t_{y}$ is the track identity assigned to anchor $y$, $D_{ab}$ is the non-squared Euclidean distance between the embeddings of anchor $a$ and anchor $b$ and $m$ is the margin (set to $m=0.1$ in experiments). Thus triplets are produced by finding a hard positive and a hard negative for each anchor. In practice, we  sample 64 triplets for computing the loss.

For detection losses we follow a target assignment convention similar to that described in~\cite{lin2017focal}. Specifically, an anchor is assigned to a groundtruth box if it has intersection-over-union (IOU) overlap of 0.5
or higher and to background otherwise.  Additionally, for each groundtruth box, we force the nearest anchor (with respect to IOU) to be a match even if this IOU is less than the threshold.  
For triplet losses, we follow a similar convention for assigning track identities to anchors, using a more stringent criterion of $IOU=.7$ or higher for positive matches --- finding that this more stringent criterion leads to improved tracking results. Only anchors that match to track identities are used to produce triplets. Further, triplets are always produced from within the same clip. 

We train on Google TPUs (v3)~\cite{kumar2019scale} using Momentum SGD with weight decay 0.0004 and momentum 0.9.  We construct each batch using 128 clips, drawing two frames for each clip spaced 8 frames apart (Waymo sequences run at 10Hz, so this corresponds to a temporal stride of 0.8 seconds).  Batches are placed on 32 TPU cores, colocating frames from the same clip, yielding a per-core batch size of 4 frame pairs. 
Unless otherwise specified, images are resized to
 $1024\times 1024$ resolution, and in order to fit this resolution in TPU memory, 
we use mixed precision training with bfloat16
type in all our training runs~\cite{wang2019}.

We initialize the model using a \modelname model (removing embedding projections) pretrained on the COCO dataset.  Next (unless otherwise stated)
we train using a linear learning rate warmup for the first 1000 steps increasing to a base learning rate of 0.001, then use a cosine annealed learning rate~\cite{loshchilov2016sgdr} 
for 9K steps. Following RetinaNet, we use random horizontal flip and random crop data augmentations. We also allow all batch norm layers to update independently during training and do not force them to be tied even if neighboring convolution layers are shared.

\vspace{-1mm}
\subsection{Inference and Tracking Logic}\label{sec:trackingsystem}
\vspace{-1mm}

\looseness -1 We use our embeddings within a simple single hypothesis tracking system
based on greedy bipartite matching.
At inference time we construct a \emph{track store} holding stateful track information.  For each track we save previous detections (including bounding boxes, class predictions and scores), embedding vectors and 
``track states'' indicating whether a track is alive or dead (for simplicity, we do not consider tracks to ever be in a ``tentative'' state, c.f.~\cite{wojke2017simple}).   
We initialize the track store to be empty, then for each frame in a clip, we take the embedding vectors corresponding to the top scoring 100 detections from \modelname.

These detections are filtered by score thresholding and then we compare the surviving embedding vectors against those in the track store via some specified similarity function $S$ and run greedy bipartite matching disallowing matches where the cosine distance is above a threshold $1-\epsilon$. Based on this greedy matching, we then add a detection to an existing track in the track store or we use it to initialize a new track.  In our experiments, our similarity
function $S$ is always a uniformly weighted sum of IOU overlap (using a truncation
threshold of 0.4) and a cosine distance between embeddings.

For each live track in the track store, we save up to $H$ of its most recent (detection, embedding vector, state) triplets thus allowing new detections to match to any of these $H$ most recent observations for all tracks. Tracks are kept alive for up to 40 frames for re-identification purposes.  Conversely we mark a track as dead if it has not been re-identified in over 40 frames.

%% file: experiments.tex
\begin{figure*}[t!]
    \centering\small
    \begin{tabular}{c|c|c|c|c|c|c}
        Architecture & Share task weights & $m_1$ & $m_2$ & $K$ & mAP & Inference time (ms per frame) \\
        \hline
        RetinaNet & No & - & - & 6 & 36.17 & 45 \\
        RetinaNet & Yes & - & - & 6 & 35.35 & 40 \\
        RetinaNet & No & - & - & 1 & 31.45  & 37 \\
        RetinaNet & Yes & - & - & 1 & 30.71 & 30 \\
        \modelname & - & 1 & 3 & 6 & 35.11 & 83 \\
        \modelname & - & 2 & 2 & 6 & 35.55 & 75 \\
        \modelname & - & 3 & 1 & 6 & 35.74 & 74 \\
    \end{tabular}\vspace{-2mm}
    
    \caption{\footnotesize  \textbf{COCO17 ablations.} Performance of vanilla RetinaNet and
\modelname (without tracking embedding layers) in terms of single 
image object detection performance on COCO17.  $m_1$ denotes the number of
task-shared post-FPN layers and $m_2$ denotes the number of task-specific
post-FPN layers.
}\vspace{-3mm}
    \label{tab:per_anchor_ablations_coco}
\end{figure*}

In our experiments we  focus on the recently released
Waymo Open dataset~\cite{waymo_open_dataset} v1 (\emph{Waymo} for short).
We also report results
on the larger v1.1 release in Section~\ref{sec:waymov11}.
This dataset
contains 
annotations on 200K frames collected at 10 Hz in Waymo vehicles and 
covers various geographies and weather conditions.  
Frames come from 5 camera positions (front and sides).
For the purposes of this paper,
we focus on 2d detection and tracking  and more specifically only
on the `vehicle' class as the dataset has major class imbalance,
which is not our main focus. 
In addition to Waymo, we report ablations
on the COCO17 dataset~\cite{lin2014microsoft}.

Finally we evaluate both detection and tracking metrics as measured by 
standard mean AP~\cite{everingham2015pascal,lin2014microsoft,girdhar2018detect} (mAP) as well as
CLEAR MOT tracking metrics~\cite{bernardin2008evaluating,milan2016mot16}, specifically using the COCO AP (averaging over IOU thresholds between 0.5 and 0.95) and the py-motmetrics library.\footnote{\url{https://github.com/cheind/py-motmetrics}}
We also benchmark using Nvidia V100 GPUs reporting inference time 
in milliseconds per frame.  For all models we only benchmark the ``deep learning part'',
ignoring any bookkeeping logic required by the tracker
which is typically very lightweight.

Evaluating a model simultaneously for detection and tracking requires some care.
Detection mAP measures a model's average ability to trade off 
between precision and recall without requiring a hard operating point 
 --- it's therefore better to use a low or zero score threshold for detection mAP.
However CLEAR MOT tracking metrics such as MOTA  require selecting a
single operating point as they directly reference true/false positives and in practice are fairly sensitive to these hyperparameter choices.
It is often better to use a higher
 score threshold to report tracking metrics so as to not admit too many false positives.  
In our experiments we simple use separate thresholds for evaluation: we evaluate our model as a detector using a near-zero 
score threshold and as a tracker using a higher score threshold.

\begin{figure*}
    \centering\small
    \begin{tabular}{c|c|c|c|c|c|c|c|c}
        Architecture &  Share task weights & $m_1$ & $m_2$ & $m_3$ & $K$ & MOTA & mAP & Inference time (ms per frame) \\
        \hline
        RetinaNet & No & - & - & - & 6   & -         & 38.19 & 34 \\
        RetinaNet$^*$ & No & - & - & - & 6   & 38.02         & 37.43  & 44 \\
        RetinaNet & Yes & - & - & - & 6  & -         & 37.95 & 30 \\
        RetinaNet$^*$ & Yes & - & - & - & 6  & 37.63         & 36.75 & 40 \\
        RetinaNet & No & - & - & 2 & 1   & 30.94     & 35.20 & 33 \\
        RetinaNet & Yes & - & - & 2 & 1  & 31.20     & 35.08 & 29 \\
        \modelname & - & 1 & 3 & 2 & 6 & 38.71     & 37.96 & 88 \\
        \modelname & - & 2 & 2 & 2 & 6 & 39.08     & 38.14 & 81 \\
        \modelname & - & 3 & 1 & 2 & 6 & 39.12     & 38.24 & 70
    \end{tabular}\vspace{-2mm}
    \caption{\footnotesize  \textbf{Waymo ablations.}  Performance of vanilla RetinaNet and
\modelname (including tracking embedding layers) in terms of detection mAP and tracking MOTA on the Waymo Open Dataset.  $m_1$ denotes the number of
task-shared post-FPN layers, $m_2$ denotes the number of task-specific
post-FPN layers, and $m_3$ denotes the number of embedding layers. 
RetinaNet$^*$ is a vanilla RetinaNet model (with $K=6$) trained with tracking
losses where instance embedding vectors are shared among ``colliding anchors''.
}\vspace{-3mm}
    \label{tab:waymo_ablations}
\end{figure*}

\vspace{-2mm}
\begin{figure}
    \centering\small
    \begin{tabular}{c|c|c}
        \# embedding layers & MOTA & mAP  \\
        \hline
        0 & 38.52 & 37.93 \\
        2 & 39.19 & 38.24  \\
        4 & 38.85 & 38.24  
    \end{tabular}\vspace{-2mm}
    \caption{\footnotesize  Track embedding subnetwork
    depth ablation. 
    We train versions of \modelname with $m_3=0$, $2$, and $4$ projection layers.}\vspace{-3mm}
    \label{fig:depth_ablation}
\end{figure}

\subsection{Evaluating \modelname as a  detector}
\vspace{-1mm}

As a preliminary ablation (Table~\ref{tab:per_anchor_ablations_coco}), 
we study the effect of our architectural 
modifications to RetinaNet on standard single image detection by 
evaluating on COCO17.  In these experiments we drop the embedding layers
of \modelname since COCO is not a video dataset.

For these experiments only, 
we train with a slightly different setup compared to our later
Waymo experiments.  We use Resnet-50 as a base feature extractor
(Imagenet initialized),
and train at $896\times 896$ resolution with 
bfloat16~\cite{wang2019} mixed precision. 
We train with batches of size 64 split across 8 TPU v3 cores,
and performing per-core batch normalization.
We use a linear learning rate warmup for the first 2K steps increasing to a base learning rate of 0.004, then use a cosine annealed learning rate~\cite{loshchilov2016sgdr} 
for 23K steps.
Note that we could use 
heavier feature extractors or higher image resolutions 
to improve performance, but the main objective of these ablations is to shed
light on variations of the Post-FPN subnetworks of RetinaNet and
\modelname.

Recall that $m_1$ and $m_2$ refer to the number of convolutions 
 for the task-shared and task-specific post-FPN subnetworks
respectively.  
We set $m_1+m_2=4$ so as to be comparable to RetinaNet.  $K$ is the number of anchor
shapes per location which we set to 6 by default but to show that
having multiple anchor shapes per location is important for detection, we also
compare against
a simplified RetinaNet which uses only 1 box per location.
Finally we experiment with a version of vanilla RetinaNet 
where the task-specific subnetworks are forced to share their
weights (the ``Share task weights'' column in Table~\ref{tab:per_anchor_ablations_coco}) since this is
closer to the task-shared post-FPN layers of  \modelname.

\looseness -1 We note firstly that using $K=6$ anchors per location is very important
to strong performance on COCO and that it is better to have separate
task-specific subnetworks than it is to share, confirming observations
by~\cite{lin2017focal}.  
We also observe that by using \modelname, we are able to 
extract per-instance features by design
(which we will next use for tracking, but could be generally useful)
while achieving similar detection performance on COCO.
If one does not need per-instance level features, one can still
get slightly better numbers with the original prediction head
layout of RetinaNet (which is similar 
to that of SSD~\cite{liu2016ssd} 
and the RPN used by many papers, e.g.,~\cite{ren2015faster,he2017mask}). 
Among the 3 settings of $(m_1, m_2)$ for \modelname, 
we find that using 3 task-shared layers ($m_1=3$) followed a single 
 task-specific layer ($m_2=1$), has a slight edge over the other
 configurations.

We report running times (averaged over 500 COCO  images)
in Table~\ref{tab:per_anchor_ablations_coco}.
Our modifications increase running time over vanilla RetinaNet --- this is
unsurprising since the cost of the post-FPN subnetworks have now been
multiplied by $K$.  Among the three variants
of \modelname, $(m_1=3, m_2=1)$ is again the fastest.

\begin{figure*}
    \centering\small
    \begin{tabular}{c|c|c|c|c|c|c|c|c|c}
        Model & MOTA & TP  & FP  & ID switches & mAP & Inference time (ms per frame) \\
        \hline
        Tracktor &   35.30 & 106006 & 15617 & 16652 & 36.17  & 45 \\
        Tracktor++ & 37.94 & 112801 & 15642 & 10370 & 36.17 & 2645 \\
        \modelname & 39.19 & 112025 & 11669 & 5712 & 38.24 & 70 
    \end{tabular}\vspace{-2mm}
    \caption{\footnotesize We compare \modelname to
Tracktor/Tracktor++~\cite{bergmann2019tracking}
which are currently state of the art on the 
MOT17 Challenge. %
}\vspace{-2mm}
    \label{tab:vs_tracktor}
\end{figure*}

\begin{figure}[t!]
    \centering\small
    \begin{tabular}{c|c|c|c}
        Model & MOTA & mAP & Inference \\
         &  &  &  time (ms) \\
        \hline
        IOU baseline & 35.36 & 38.53 & 70 \\
        \modelname w/o triplet loss & 37.92 & 38.58 & 70 \\
        \modelname, w/R-50 ReID & 37.39 & 38.58 & 80 \\
        \modelname & 39.19 & 38.24 & 70 
    \end{tabular}\vspace{-2mm}
    \caption{\footnotesize Comparison of joint training (\modelname) with alternatives:
    (1) IOU based similarity tracker, (2) \modelname w/o triplet loss,
    (3) \modelname w/R-50 ReID, 
    }\vspace{-3mm}
    \label{fig:joint}
\end{figure}

\vspace{-1mm}
\subsection{Architectural ablations}
\vspace{-1mm}

For our remaining experiments we evaluate on  Waymo,
this time including the embedding network with triplet loss training
and additionally evaluating tracking performance
using the system described in Section~\ref{sec:trackingsystem}.

We first ablate the depth of the embedding network (see Table~\ref{fig:depth_ablation})
in which we train models using $m_3=0$, 2 and 4 projection layers (fixing $m_1=3$ and $m_2=1$
as was shown to be best on the COCO ablation above), obtaining best
performance for both detection and tracking with 2 layers.

Setting $m_3=2$ layers for the embedding subnetwork, we present our ablations on the
Waymo dataset in Table~\ref{tab:waymo_ablations}, training
via the method described in Section~\ref{sec:training}.

To demonstrate the value of \modelname's
anchor-level features for tracking,
we evaluate two baseline versions of the vanilla RetinaNet architecture ---
(1) one where we use $K=1$ anchor shapes since in this case it is possible to extract 
per-instance feature vectors, and (2) the standard $K=6$ setting where during tracking
we simply force embeddings for anchors that ``collide'' at the same spatial
center to be the same (we refer to this baseline as RetinaNet$^*$).

As with the COCO ablations, we see that using multiple
($K=6$) anchor shapes is important to both detection
and tracking metrics.
Thus it is unsurprising that
\modelname significantly outperforms the RetinaNet based ($K=1$) tracking baseline likely mostly by virtue of being a stronger detector.
However both RetinaNet$^*$ rows exhibit lower MOTA and mAP results compared to their
non-starred counterparts, suggesting that ``abusing'' vanilla RetinaNet to perform
tracking by ignoring colliding anchors is harmful both for detection and tracking, 
thus underscoring the importance of \modelname's per-anchor embeddings.

Our best \modelname configuration reaches 39.12 MOTA and has a mAP of 38.24.
In contrast to the COCO ablations where vanilla RetinaNet retains a
slight edge over \modelname, here 
we see that \modelname outperforms  RetinaNet as a detector, suggesting that by
 including  tracking losses, we are able to boost detection performance.

Finally with a running time of 70ms per frame, we note that
inference with \modelname is faster than
the sensor framerate (10 Hz) in the Waymo dataset. 
Compared to the COCO setting, \modelname must run additional
convolution layers for embeddings, but since COCO has 80 classes which
makes the top of the network slightly heavier, the final running time is
slightly lower in the Waymo setting.

\vspace{-1mm}
\subsection{Joint vs Independent training}\label{sec:joint}
\vspace{-1mm}

To demonstrate the benefit of joint training with detection and 
tracking tasks, we now compare \modelname against three natural baselines
which use the same tracking system as \modelname but change the underlying
data association similarity function
(Table~\ref{fig:joint}):
\vspace{-2mm}
\begin{itemize}\denselist
    \item An \emph{IOU baseline}, where  detection similarity
is measured only by IOU overlap  (with no embeddings),
    \item \emph{\modelname w/o triplet loss}, in which we ignore
the triplet loss (and thus do not train the model specifically for tracking) 
and measure
embedding similarity via the per-instance
feature vectors $F_{i,k}$, and
    \item \emph{\modelname w/R-50 ReID}, in which again we ignore
triplet loss when training \modelname and feed the detections
to an offline-trained re-identification (ReID) model. For the ReID model, we train a Resnet-50 based TriNet model~\cite{hermans2017defense} 
to perform ReID on Waymo.
\end{itemize}
We observe that even the IOU-only  tracker provides
a reasonably strong baseline on  Waymo, most likely
by virtue of have a strong detection model --- it is likely
that this tracker is more accurate when the car is driving slowly
(compared to, e.g., highway driving).  However, using visual embeddings allows us to outperform this simple baseline in all cases,
and \modelname when trained with detection and metric learning
losses jointly outperforms these baselines.

\vspace{-2mm}
\subsection{Comparison against state of the art}\label{sec:waymov11}
\vspace{-2mm}

We finally compare (Table~\ref{tab:vs_tracktor}) 
against the recent Tracktor and Tracktor++ algorithms
which are currently state of the art on MOT Challenge. 
For these experiments we use our own Tensorflow reimplementations
of Tracktor and Tracktor++ which adds a ReID component and 
camera motion compensation (CMC). 
Our implementation differs in some details
from that described in the original paper in that it is based on the
Tensorflow Object Detection API~\cite{huang2017speed}
and does not use an FPN.  
We use the same ReID model as the one in Section~\ref{sec:joint}, 
which matches the approach taken in the Tracktor paper.
To verify that our reimplementations are
competitive, we submitted results
from our Resnet-101
based Tracktor models to the official MOT Challenge server, which 
achieve nearly identical MOTA numbers as the official submission which 
uses an FPN (53.4 vs. 53.5).  We also submitted results from
a Resnet-152 based Tracktor which currently outperforms all 
entries on the public leaderboard (with 56.7 MOTA).  

On  Waymo, we use a  Resnet-50 based Tracktor  running at 
$1024\times 1024$ resolution to be comparable to our model.
If we compare the  Tracktor (without
CMC or ReID) MOTA score to the IOU tracking performance in Table~\ref{fig:joint},
we see that the two approaches are roughly on par.  We believe that IOU based tracking can achieve parity with 
Tracktor here due to (1) having highly accurate detections to begin with,
and (2) significant camera motion which hurts  Tracktor.

In fact we observe that
Tracktor needs the `++` to significantly outperform the IOU based tracker.
However it is far slower --- in addition to running
Faster R-CNN, it must run a second Resnet-50 model for ReID
followed by CMC (which is time consuming).\footnote{
To benchmark the runtime of CMC on Waymo, we use the same function used
by the authors of~\cite{bergmann2019tracking} 
(OpenCV's \emph{findTransformECC} function with `MOTION\_EUCLIDEAN` option),
and run on a workstation with 56 Intel(R) Xeon(R)  E5-2690 v4 2.60GHz CPUs (w/14 cores/CPU).
}

\modelname outperforms both variants on tracking and detection. It 
is able to achieve these improvements by significantly reducing the number of
false positives and ID switches.  And despite being slower than  vanilla
Tracktor  (whose running time is dominated by Faster R-CNN), \modelname is 
significantly faster than Tracktor++.

\paragraph{Evaluation on the Waymo v1.1 dataset.}
As a baseline for future comparisons, we also reproduce our evaluations on the 
Waymo v1.1 release with $\sim 800$K frames for
training containing $\sim 1.7$M annotated vehicles.
For these evaluations, we train for 100K steps with a base learning rate of 0.004 
(and all other hyperparameters fixed).
Results are shown in Table~\ref{fig:waymo1.1}, where we again see the same
trends with \modelname significantly outperforming a baseline IOU based tracker as
well as outperforming Tracktor++  
with a significantly faster running time.

\begin{figure}[t!]
    \centering\small
    \begin{tabular}{c|c|c|c}
        Model & MOTA & mAP & Inference \\
         &  &  &  time (ms) \\
        \hline
        IOU baseline & 38.25  & 45.78 & 70 \\
        Tracktor++ & 42.62 & 42.41 & 2645 \\
        \modelname & 44.92 & 45.70 &  70  
    \end{tabular}\vspace{-2mm}
    \caption{\footnotesize Evaluations on the Waymo v1.1 dataset (which has a $4\times$
    larger training set than the v1 dataset).
    }\vspace{-3mm}
    \label{fig:waymo1.1}
\end{figure}

%% file: conclusion.tex
In this paper we have presented a simple but effective model,
\modelname, which trains jointly on detection and tracking tasks and
extends single stage detectors to handle instance-level attributes,
which we note
may  be of independent interest for  applications beyond tracking.

Additionally we have demonstrated the effectiveness of joint training over 
the prevailing approach of training independent
detection and tracking models.  This approach allows \modelname
to outperform the current state of the art in multi-object tracking 
while being significantly faster and 
able to track through long periods of object disappearance.
Finally we hope that our work can serve as a strong baseline for
future research in detection and tracking.